\documentclass[journal,final]{IEEEtran}
\usepackage{cite}

\ifCLASSINFOpdf
  \usepackage{graphicx}
\else
\fi
\usepackage[cmex10]{amsmath}
\usepackage{array}
\usepackage{url}

\usepackage{threeparttable}

\begin{document}
%
\title{Predicting the Effectiveness of Self-Training: Application to Sentiment Classification}
%
%
%

\author{Vincent~Van~Asch,%
        Walter~Daelemans
\thanks{Both authors are with CLiPS, University of Antwerp, Belgium}%
\thanks{For contact information, see www.clips.uantwerpen.be}}%
\maketitle

\begin{abstract}
The goal of this paper is to investigate the connection between the
performance gain that can be obtained by self-training and the
similarity between the corpora used in this approach.  Self-training is a
semi-supervised technique designed to increase the performance of machine
learning algorithms by automatically classifying instances of a task and
adding these as additional training material to the same classifier.
In the context of language processing tasks, this training material is
mostly an (annotated) corpus. Unfortunately self-training does not always lead to a
performance increase and whether it will is largely unpredictable. We
show that the similarity between
corpora can be used to identify those setups for which
self-training can be beneficial. We consider this research as a step
in the process of developing a classifier that is able to adapt itself
to each new test corpus that it is presented with.
\end{abstract}


%
\IEEEpeerreviewmaketitle

\section{Introduction}
%
%
%
%
\IEEEPARstart{W}{hen} developing and testing techniques to increase
the performance of natural language processing systems, the choice of
the corpora to test the technique may influence the efficacy of the new
technique. For this reason it is important to introduce a second dimension, 
apart from the performance scores, that can describe the corpora that 
have been selected. We have chosen to introduce similarity scores in a self-training
setup in order to identify those setups for which the self-training technique is useful.   


\subsection{Self-training procedure}
\label{sec:procedure}

Self-training is a semi-supervised machine learning method developed
mainly to enhance the performance of a machine learner on corpora that
are more dissimilar from the training corpus than one would prefer.
When train and test data distributions are too different, models
trained on the training data will not generalize to the test data.

The self-training procedure is, in its plain form, a two-step
technique. Variants of the self-training procedure include an instance
selection step to increase the probability of adding only informative
instances to the training data. We do not address instance selection in this paper.

Three corpora are needed for self-training: a labeled training corpus, a labeled test corpus, and an unlabeled additional corpus \cite{Charniak97}. During self-training, a model is learned from the training data and it is applied to the unlabeled data.
Thus, the additional training data is created by automatically labeling unlabeled data. Next, the (partially incorrectly) labeled additional data is appended to the original training data (\emph{self-training step 1}). 
This first labeling step is followed by a second training phase using
the original training data plus the newly labeled data. The model
resulting from this phase is then used to label the test data
(\emph{self-training step 2}). The expectation is that labeling the
test data in \emph{self-training step 2} yields more correct labels
than simply labeling the test data depending only on the information
present in the original training data.

It remains controversial whether self-training is a useful method; it
is not shown to lead to performance gain for every experimental setup.
Reference \cite{Sagae10}  argues that self-training is only beneficial in
those situations for which the training and test data are sufficiently
dissimilar, but other factors -- most obviously labeling accuracy of the unlabeled data -- may have an influence too. Rather than dismissing the self-training technique as dysfunctional, we want to identify those setups for which performance gain can be expected.

Since self-training is a technique to enhance the performance in
situations of data sparseness, it can be linked to the notion of
domain adaptation. For an introduction to domain adaptation see \cite{Daume06}. In a domain adaptation context the test corpus and the training corpus originate from different domains. Unfortunately, the term \emph{domain} is a vague concept and defining a domain in an objective manner is a contentious task. Reference \cite{DLee01}, drawing upon \cite{Biber88} and the EAGLES initiative \cite{Sinclair96}, mentions two types of parameters to categorize corpora: external (intended audience, purpose, and setting) and internal (lexical or grammatical (co)occurrence features). A domain would then be an external parameter, \emph{viz.}\ the subject field, like finances, mathematics, or linguistics.

However, in machine learning, the only requirement of a domain is that
the underlying distribution of the words of a corpus coming from that
domain is different from the distribution of other domains. This
broadens the scope of a domain, but it still remains unclear how to
express the difference between the distributions and, as a
consequence, there are no objective boundaries to a domain. In
practice, domain corpora are straightforwardly gathered from different
subject fields, but it would be equally valid to collect texts using different registers, like newswire
data, e-mail correspondence and novels, all on the same subject.
Irrespective of how domain corpora are collected, the basic assumption
is that cross-domain machine learning will suffer from sub-optimal
generalization.

The objective of this study is to investigate the foundations of a
self-adapting classifier. Consider a system that has access to one
labeled training corpus and a collection of unlabeled corpora coming
from varying domains. If an unseen, unlabeled text is presented to
this system, this system should be able to select from the unlabeled
corpora the corpus (or corpora) that would facilitate the labeling
process most. In this manner, the classifier would be able to adapt itself to each newly presented text. In this paper, two assumptions of this scenario are tested: the assumption that it is possible to select an unlabeled corpus that increases the performance and the assumption that it is better to select an unlabeled corpus than it is to include all additional unlabeled data. The first assumption corresponds to the assumption that self-training can be helpful if a suitable combination of corpora is involved. 

To be able to select the best-suited unlabeled corpus, the classifier in this study relies on similarity measures. A similarity measure expresses the similarity of two corpora and various implementations are available. Common implementations are unknown word ratio and Kullback-Leibler (see Section~\ref{sec:simmeas}). Combining the concept of self-training and domain similarities may enable us to make a distinction between useful and harmful self-training setups. 

The bulk of his paper tackles the question when self-training is helpful. It starts with an overview of related research (Section~\ref{sec:rs}) followed by an overview of the different elements that are used in the experiments (Section~\ref{sec:ede}), experimental results (Section~\ref{sec:exps}), discussion and additional experiments (Section~\ref{sec:dis}), and conclusions (Section~\ref{sec:con}).

\section {Related research}
\label{sec:rs}
 The concepts of self-training were first introduced by \cite{Charniak97}, but more recent examples are given by \cite{Jiang07}, \cite{McClosky10}, and \cite{Sagae10}. There are many variations on self-training. Reference \cite{Dong2011} combines the self-training concept with ensemble learning for citation classification. Thus, instead of using a single classifier that has to be trained, they use an ensemble of classifiers. Another interesting approach is the combination of self-training with active learning \cite{Liu2013}. Instead of using the entire training set to label the unlabeled data, a portion is kept apart. This held-out data is labeled together with the unlabeled data in \emph{self-training step 1}. The most confidently classified instances of the unlabeled data are added to the training corpus together with the least confidently labeled held-out data. The intention is to select useful data in a more rigorous manner.

It is possible to use similarities between corpora as such, using
their outcome to draw inferences about them \cite{Verspoor09,Biber10},
but an additional interesting usage involves applying similarities in a machine learning setup.

Similarities are used in natural language processing in various situations ranging from feature selection \cite{Pietra97,Mitra02} and measuring the similarity between two language models \cite{Lee01,Gao02} to training corpus creation \cite{Chen09,McClosky10,Moore2010,Plank11}. An example of training corpus creation is presented by \cite{Mansour09}. They employ the R\'enyi divergence \cite{Renyi61} to create a combination of different training corpora in order to increase the labeling performance for a specific test corpus.

Similarities have also been used to predict the performance of machine learners \cite{Zhang09,Asch10}. A good example of such an application is the prediction of parsing accuracy \cite{Ravi08}.
Although there are issues when similarity measures are used (see Section~\ref{sec:issues}), good results have been obtained for these tasks. Apart from the R\'enyi divergence, some of the similarities that are used are perplexity, Kullback-Leibler divergence \cite{Kullback51}, and the Skew divergence \cite{Lee99}.

Despite the fact that authors have shown that a similarity \cite{Asch10,Plank11} or a linear combination of similarities \cite{McClosky10} can be successfully used to link the similarity between domains to the performance of a natural language processing system, no consensus exists about which similarity or combination of similarities is best suited for the task. The best similarity is not selected on theoretical grounds but by testing a range of similarities and selecting the best one. For the research of\cite{Yuan2005}, measuring the cross-entropy between two domains offered the best results when adapting a baseline language model to a new domain.

The bag-of-words approaches presented above are not the only manner of computing similarities automatically. It is possible to obtain a richer comparison of texts using the semantics. Current research focuses on semantic textual similarity (STS) \cite{Agirre2013}. Most of these similarities draw from various sources like dependency parsing, part-of-speech tagging and/or latent Dirichlet allocation. An interesting software package in this context is the DKPro Similarity package \cite{Bar2013}, which implements various semantic similarities in addition to less complex string matching similarities.

Similarity measures have been used on a range of corpora and currently we know of two papers that carry out domain similarity research on the same corpus that we use \cite{Ponomareva12,Remus12}.


Reference \cite{Ponomareva12} shows that instead of exploiting the correlation
between the test/training similarity and the accuracy, one could also
use the correlation between the similarity and the accuracy drop. The
accuracy drop is the difference between the accuracy of an in-domain
experiment, using the test corpus, and the accuracy of a cross-domain
experiment. In this setup, the in-domain accuracy is considered to be
an upper bound. Reference \cite{Ponomareva12} also introduces the notion of domain complexity, which may be expressed by the percentage of \emph{rare} words in a domain. Examining the corpus of \cite{Blitzer07}, they observe that less complex source domains tend to give a smaller accuracy drop on more complex target domains. They prefer to use $\chi^2$ in combination with inverse document frequencies (IDF) to measure domain similarity.  

Reference \cite{Remus12} investigates instance selection for the corpus of \cite{Blitzer07}. He strives for the identification of the most helpful instances using the Jensen-Shannon divergence.

\section{Experimental design}
\label{sec:ede}

\subsection{Definitions}

There are two levels of evaluation in this paper. The first level is the
self-training experiment and the second level is the evaluation of how
well self-training gain can be predicted. For maximal clarity, some explicit definitions are given here first.

\textbf{similarity} A number expressing the degree to which two corpora are similar. The higher this number, the less similar the corpora are.\footnote{Because higher values express less similarity this is sometimes called \emph{distance}. However, the term distance entails certain mathematical properties that we do not demand from a similarity measure.} The exact meaning of \emph{similar} depends on the similarity measure that is used.\\
\indent \textbf{labeling experiment} A labeling experiment consists of labeling instances. Often this involves a training phase using a labeled training corpus and a test phase during which the unlabeled instances of the test corpus are assigned a label. This is a standard classification experiment. We use the term \emph{labeling} to be able to differentiate this experiment from the second level classification experiment that involves the classification of self-training setups according to the self-training gain. \\
\indent \textbf{labeling performance} The performance of a labeling experiment. Labeling performance can be quantified using accuracy, precision, recall, F-score, etc. and these scores are calculated using the gold standard of the test data.
\\ \indent \textbf{self-training experiment} Each self-training experiment is linked to a particular setup involving three corpora: a labeled training corpus, a labeled test corpus and unlabeled additional data. This is an experiment as described in Section~\ref{sec:procedure}.
\\ \indent \textbf{self-training gain} The performance gain obtained when the labeling performance for a given test and training set is compared with and without the introduction of self-training. The inverse is self-training loss.
\\ \indent\textbf{self-training gain prediction} Apart from the labeling experiments, which are the basis of the self-training experiments, this is the second type of classification experiment discussed in this paper. It can be regarded as a \emph{second level} experiment, meaning that first a range of self-training experiments is conducted before self-training gain is predicted. The classification  consists of separating the setups that lead to self-training gain from the others.
\\ \indent \textbf{prediction performance} The performance of self-training gain prediction. A self-training experiment counts as a true positive if it is correctly predicted to result in self-training gain. Prediction performance can be quantified using accuracy, precision, recall, F-score, etc.

\subsection{Corpus and labeling task}

Self-training can be applied to every supervised machine learning problem. In this paper, the labeling task consists of a binary sentiment classification task. The goal is to label an instance based on a product review according to the sentiment expressed in the review: Is the review favorable to the product or not? 

The instances for this task are bag-of-word instances coming from the sentiment classification corpus of \cite{Blitzer07}.\footnote{Multi-Domain Sentiment Dataset (v.\ 2.0) in April 2013 retrieved from\\ \url{www.cs.jhu.edu/~mdredze/datasets/sentiment/unprocessed.tar.gz} .} The main reason why we chose this corpus is that it contains data for various domains which is labeled for the same task.

To minimize corpus size effects, the corpora for the different domains are normalized to a size of 2,500 instances. We chose this number as a trade-off between sufficient corpus size and sufficient number of domains that have more than 2,500 instances to sample from.   
In the end, 13 domains meet the corpus size constraints: \emph{beauty, baby, camera \& photo, sports \& outdoors, health \& personal care, apparel, toys \& games, video, kitchen \& housewares, electronics, dvd, books, and music}.

After randomly sampling the instances, a script is used to convert the instances into a format fit for the machine learner, i.e.\ SVMLight v6.02 \cite{Joachims99}. The machine learner is used with default settings.

For each self-training experiment, three different corpora are needed. With 13 domains, a total of 1,716 distinct setups ($= 13 \times 12 \times 11$) are conceivable. Of these 1,716 self-training experiments, 94\% of the setups lead to performance loss. It is clear that with only 106 setups in the positive class, self-training -- with little additional data -- is more often detrimental to performance than it is helpful. This illustrates that being able to identify setups leading to decreased accuracy can be of help when the use of  self-training is considered.





We briefly discuss a few general observations involving the data. When carrying out regular cross-domain labeling experiments, the average macro-averaged F1-score is 60.61\%, equalling an accuracy of 85.44\%. Although the experiments are not directly comparable, the cross-domain accuracy is in the same region as reported by \cite{Blitzer07} meaning that the machine learner is not underperforming. 

Carrying out the 1,716 self-training experiments results in an average F1-score loss of 3.5\%. If performance increases, an average of 0.6\% is added to the score. 
The differences in F1-score are rather small, but for the best self-training setups the difference is statistically significant at the 1\% confidence level.\footnote{Using approximate randomization testing \cite{Noreen89,Yeh00}.} 
Also note that in a regular self-training setup, considerably more additional unlabeled data is added, which may lead to more self-training gain. We conducted a small experiment with dvd (training), video (test), and apparel as additional data. Using all apparel data (8,940 instances instead of 2,500) leads to a labeling performance F-score increase of 4.98\% instead of 4.29\%. This small experiment shows that adding more additional data, as one would normally do, may indeed lead to a higher self-training gain. Although the self-training gain in our experiments is rather small, more data may lead to more gain thus making the selection of the right corpora more relevant.

\subsection{Similarity measures}
\label{sec:simmeas}

An important issue is how the similarity between corpora can be measured. The features of the corpus of \cite{Blitzer07} consist of tokens. 
An instance can be considered a bag-of-words and can be converted into a vector. The values in the vector indicate whether a given token occurs in the sample text or not. In this manner, an entire corpus can be reduced to a single vector, namely the centroid of all instance-based vectors in that corpus. If we have a vector for the test corpus and one for the training corpus, for example, cosine similarity between the two domain vectors can be computed. During the experiments, the cosine similarity and the Euclidean distance between the two vectors are computed. To make the similarity independent of sample size, the actual values in the corpus-based vectors are not the raw counts but the point-wise mutual information (pmi) values. Point-wise mutual information also smoothes down the influence of large token count differences. Given the two raw count vectors of the two corpora that are compared, the pmi-value of token $t$ in vector $v$ becomes: 
\begin{align}
pmi(t,v) &= log \big ( n \frac{c^{v}_{t}}{c^{*}_{t}c^{v}_{*}}  \big) \\
c^{v}_{t} &: \text{count of token $t$ in vector $v$} \nonumber \\
c^{*}_{t} &: \text{sum of all counts of token $t$} \nonumber \\
c^{c}_{*} &: \text{sum of the counts in vector $v$} \nonumber \\
n &: \text{sum of all counts} \nonumber
\end{align}

Instead of calculating a distance between vectors, it is also possible to consider the vector as a probability distribution of tokens occurring in a corpus. Similarity between probability distributions can be calculated with e.g the Kullback-Leibler divergence (KL; \cite{Kullback51}):

\begin{equation} KL(P; Q) = \sum_k p_k log_2\Big( \frac{p_k}{q_k}\Big) \end{equation}

\noindent with $p_k$ the value in the vector of the test corpus $P$ at position $k$ and $q_k$ the value of the vector of the training corpus $Q$ at position $k$.\footnote{When $q_k=0$, smoothing is applied by setting $q_k = 2^{-52}$.}   

Apart from the Kullback-Leibler divergence, we also implemented the Jensen-Shannon divergence (JS; \cite{Lin91}):

\begin{align}
\scriptstyle{JS(P; Q)} &\scriptstyle{= \frac{1}{2}\bigg[  KL\Big (P; \frac{P + Q}{2}\Big ) + KL\Big (Q; \frac{P + Q}{2} \Big)  \bigg]}
\end{align}

\noindent Using the same notation as for the KL-divergence. The JS-divergence can be considered as a symmetric version of KL.

A fifth similarity measure that has been used is the simple Unknown Word Ratio (sUWR; \cite{Zhang09,Plank10}). The sUWR is the proportion of tokens $t$ in the test corpus $P$ that are not seen in the training corpus $Q$:

\begin{equation}sUWR(P; Q) = \frac{| \{t | t \notin Q \land t \in P \}|}{| \{ t | t \in P \} | }\end{equation}

In summary, in this study, we evaluate five similarity measures on their usefulness to predict self-training gain (i.e. their prediction performance).  

\section{Experiments}
\label{sec:exps}
\subsection{Baseline systems}

%
%
%
%
%

For our experiments, four different baselines are computed: two one-class-prediction baselines and two uncomplicated learner baselines.
All results tables contain the precision on self-training gain prediction, macro-averaged F-score for performance prediction, and the accuracy of the performance prediction.

We include the accuracy because it gives a general insight in the correct predictions. Because the majority class has more influence on the accuracy and because the accuracy ignores the precision, we also include the macro-averaged F-score. This score gives the best sense of how well a system performs.

In a practical situation, a system developer may be most interested in the precision of self-training gain prediction. Indeed, if a self-training setup is predicted to lead to a performance gain, the developer wants this prediction to be trustworthy. All other evaluation scores, like recall on gain, can be computed from the scores that are reported in this paper.\footnote{An online tool is available at \url{www.clips.uantwerpen.be/cgi-bin/vincent/scoreconverter.html}.} 

\paragraph{One-class prediction}
The baseline of the self-training gain prediction can be set to predicting that self-training will always increase labeling performance (the \texttt{POS} baseline) or that it will never increase labeling performance (the \texttt{NEG} baseline). The precision on gain, the macro-averaged F-score and the accuracy are reported in Table~\ref{tab:unsup}. As a consequence of the nature of the baselines, the accuracy of a baseline system equals the precision of the class label that is predicted.

Given the nature of the corpus, only 106 out of 1,716 setups result in self-training gain, it can be expected that always predicting a loss after self-training produces better overall scores.

\paragraph{Uncomplicated prediction}

As we will see later, self-training gain, using the sentiment prediction corpus, is highly dependent on the choice of the test and training corpus, regardless of the nature of the additional data. For this reason, using information about the outcome of previous test/training combinations will produce another type of baseline. One such baseline system may predict gain if at least one similar test/training combination in the training corpus leads to self-training gain (\texttt{ONCE}). The second may predict gain if the majority of the similar test/training instances lead to self-training gain (\texttt{MAJ}). For both baseline systems, the precision on gain, the macro-averaged F-score and the accuracy are reported in Table~\ref{tab:sup1}.

\subsection{Self-training gain prediction}

In this section, we will discuss two methods to tackle the prediction of setups that lead to gain after self-training. The unsupervised method consists of calculating an indicator based on the similarities between the corpora that are involved. The supervised method consists of training a machine learner on the similarities between the corpora. 

\subsubsection{Unsupervised}

The unsupervised way of predicting self-training gain is based on the performance indicator $\delta$ developed by \cite{Asch10}. This indicator is defined as:

\begin{equation}
\delta = \frac{\big|\frac{\text{test/training similarity}}{\text{test/additional data similarity}} - 1 \big|}{\frac{\text{test/training similarity}}{\text{test/additional data similarity}}-1} 
\label{eq:delta}
\end{equation}
 
This indicator weighs the similarity between test and training corpus relative to the similarity between test corpus and the additional data. Its design is such that $\delta$ is +1 if gain is expected from self-training; otherwise the value is -1.

\begin{table}[!t]
\renewcommand{\arraystretch}{1.3}
\centering
\begin{threeparttable}[b]
\caption{Unsupervised performance prediction.}
\label{tab:unsup}
\begin{tabular}{lrrr}
\hline
type &  precision & macro-avg. & accuracy \\
 &  on gain & F-score &  \\
\hline
\textbf{Systems} & & & \\
Cosine & 7.58 & 39.85 & 51.40 \\
Euclidean & 10.72 & 43.74 & 54.55 \\ 
KL & 6.99 & 39.13 & 50.82 \\
JS & 9.44 & 42.15 & 53.26 \\
sUWR & 7.34 & 39.56 & 51.17 \\
\emph{optimized} &&&\\
Euclidean & 17.17 & 57.69 & 82.46 \\ 
\multicolumn{2}{l}{\textbf{One-class baselines}} && \\
\texttt{NEG} & 0 & 48.41 & 93.82 \\
\texttt{POS} & 6.18 & 5.82 & 6.18 \\
\hline
\end{tabular}
\begin{tablenotes}[flushleft]
\item The one-class-prediction baselines are also given.
\item Scores are expressed as percentages.
\end{tablenotes}
\end{threeparttable}
\end{table}

For the prediction performance, all 1,716 self-training experiments are carried out and the performance is measured.
The results, substituting the five similarity measures in Equation~\ref{eq:delta}, are given in Table~\ref{tab:unsup}. 

The performance indicator $\delta$ is a simplification. As a consequence, the indicator can be optimized by varying the -1 in Equation~\ref{eq:delta}. Ideally, one would determine the required value using a separate development partition. Because of the limited data, we did not carry out these experiments. We optimized using the test partition, which leads to overfitting. Nevertheless, we report on one such system, \emph{optimized} Euclidean, in Table~\ref{tab:unsup}. The optimized value is -1.1. We include these scores because they reveal that the precision on gain remains low even after improper optimization.

From our observations, we conclude that it is unlikely to obtain reasonable prediction performance by using the performance indicator. 
The default choice is to say that self-training will never lead to gain, i.e.\ the \texttt{NEG} baseline. Only if one is anxious to discover a setup that leads to self-training gain, using the performance indicator can be helpful to at least narrow down the amount of setups that need to be tested. In this case, the Euclidean distance and the Jensen-Shannon divergence seem to be the best options to compute the similarities.

\subsubsection{Supervised}
\label{sec:sup}
For supervised performance prediction, three similarity values are taken as the features: test/train, additional/train, and test/additional.

\paragraph{Leave-one-out cross-validation}

Of the 1,716 self-training experiments, no two setups are the same. The outcome of a leave-one-out cross-validation experiment can be an estimate of how well an unseen setup can be labeled as leading to gain or loss. We choose a $k$NN-based machine learner to estimate self-training gain \footnote{TiMBL 6.4.2 -- \url{http://ilk.uvt.nl/timbl}}. The feature metric, the metric that defines how close neighbors are, is set to the Euclidean distance because the default feature metric is more appropriate for categorical features.

\begin{table}[!t]
\renewcommand{\arraystretch}{1.3}
\centering
\begin{threeparttable}[b]
\caption{Supervised performance prediction using leave-one-out cross-validation.}
\label{tab:sup1}
\begin{tabular}{lrrr}
\hline
type &  precision & macro-avg. & accuracy \\
 &  on gain & F-score &  \\
\hline
\textbf{Systems} & & & \\
Cosine & 38.10 & 66.92 & 92.37 \\
Euclidean & 49.53 & 73.22 & 93.76 \\ 
KL & 75.53 & 84.60 & 96.62 \\
JS & 71.56 & 85.36 & 96.56 \\ %
sUWR & 60.38 & 78.88 & 95.10 \\ %
\multicolumn{2}{l}{\textbf{Uncomplicated baselines}} && \\
\texttt{ONCE} & 41.90 & 77.13 & 91.43  \\
\texttt{MAJ} & 100 & 95.08 & 98.95  \\
\hline
\end{tabular}
\begin{tablenotes}[flushleft]
\item The uncomplicated-prediction baselines are also given.
\item Scores are expressed as percentages.
\end{tablenotes}
\end{threeparttable}
\end{table}

The scores are given in Table~\ref{tab:sup1}. The values in Table~\ref{tab:sup1} are better than those in the previous table. Predicting with around 70\% precision whether a setup will be profitable seems acceptable and it may convince a system developer to carry out self-training. However, there are two remarks that have to be made. Pragmatically, when designing a system, one does not always have the data to train a machine learner to predict the self-training gain. Secondly, looking at the setups that lead to self-training gain reveals a weakness of the classification. We will expand on this observation.

Each test/training/additional data combination is unique, but if only the test and training data are taken into account, 11 setups include the same test and training data. Examining which setups lead to self-training gain reveals that once a test/train pair experiences an advantage from self-training, the advantage is often present irrespective of the nature of the additional data. This means that there is information leakage using this leave-one-out cross-validation setup. Indeed, 10 test/train pairs similar to the test instance are present in the training split. Because the 10 pairs included in the training data are very likely a correct indication of the self-training gain/loss for the one pair in the test partition, the 70\% prediction precision is no surprise. Evaluation would be more relevant if the prediction performance for an entirely new setup is measured. 


Based on the results of Table~\ref{tab:sup1}, we can conclude that once it is known that carrying out self-training can increase the labeling performance, the source of the unlabeled data is less important. This observation raises questions about the nature of the differences between the test and training corpus. Performance loss due to a specific, yet unidentified, kind of difference can be mediated by adding additional information through self-training. If this difference between test and training corpus is not sufficiently prominent, self-training will be of no help. Often domain adaptation is needed without having access to knowledge about previous self-training experiments. In the next section, we will investigate what can be done best if a given test/training combination has not been seen before.

\paragraph{Tailored leave-one-out} In the previous paragraph, we have shown that knowledge about previous outcomes of a test/training combination leads to high scores. However, sometimes a given test/training combination may not have been tested yet. To examine how well self-training gain can be predicted in that situation, a tailored leave-on-out routine is implemented.

For each instance, three corpora are involved: training, test and the unlabeled data (extra). The instance contains three numbers: the train/test similarity, the train/extra similarity and the test/extra similarity. Any instance that contains any pair of corpora from the test instance is excluded from the training partition (60 instances). Also, any instance containing the same corpus as the unlabeled corpus in the test instance is excluded from the training partition (396 instances).

As a result, we have 1,716 folds with 1 instance in the test partition and 1,260 instances in the training partition. This split ensures that the similarities of the test instance are not present in the training partition.
The results are presented in Table~\ref{tab:tlo}.

\begin{table}[!t]
\renewcommand{\arraystretch}{1.3}
\centering
\begin{threeparttable}[b]
\caption{Supervised performance prediction using tailored leave-one-out cross-validation.}
\label{tab:tlo}
\begin{tabular}{lrrr}
\hline
type &  precision & macro-avg. & accuracy \\
 &  on gain & F-score &  \\
\hline
Cosine & 1.43 & 47.90 & 89.86 \\
Euclidean & 6.12 & 49.97 & 88.81 \\ 
KL & 29.63 & 60.69 & 91.90 \\
JS & 41.90 & 68.94 & 92.83 \\ %
sUWR & 2.67 & 48.38 & 89.69 \\ %
\hline
\end{tabular}
\begin{tablenotes}[flushleft]
\item Scores are expressed as percentages.
\end{tablenotes}
\end{threeparttable}
\end{table}

The scores in Table~\ref{tab:tlo} are clearly lower than the scores in Table~\ref{tab:sup1}. This could be expected because this setup emulates classifying an entirely new combination of domains. However, apart from the precision on self-training gain for the non-probability-distribution-based similarities, these scores are higher than the scores for the unsupervised method in Table~\ref{tab:unsup}. 


The experiments of this section confirm the observation that prior knowledge about the outcome of self-training experiments is the best predictor for new self-training experiments. Although this may not come as a surprising conclusion, it also holds when the tested combination of corpora was not seen before, meaning that the performance prediction classifier was able to learn from setups unrelated to the test setup. This is an indication that the similarity scores capture useful information about the corpora, in the context of sentiment classification.


\section{Discussion}
\label{sec:dis}

\subsection{Limitations of similarity scores}
\label{sec:issues}

Predicting self-training gain appears to work best if information about a collection of self-training experiments is available. Although this is a restriction on the practicability of the technique, performance prediction can be the working method of choice in selected situations. It can be useful to sum up some limitations that should be taken into account when similarity measures are used.   

\paragraph{Class label independence} Similarity measures do not use the class labels. Consider two corpora that are completely disjunct with respect to class labels but very similar in the feature space. A similarity measure will probably underestimate the difference between the two corpora and overestimate the labeling performance. Luckily there are labeling tasks (part-of-speech tagging, sentiment prediction, \ldots) for which this extreme situation is not likely to occur, but it still means that the linearity between similarity and performance should be assessed for each new task. 
\paragraph{Corpus size} Similarity measures are corpus size dependent. In our experiments all corpora are of the same size, but in a real situation this may not be the case. For example, the overlap measure quantifies the number of unseen elements in the test set given a training set. If a larger test set of the same test domain is taken, the number of unseen elements will probably decrease. The decrease is the result of chance and does not stem from a higher similarity between the domains. 
\paragraph{Similarity measure} It can be difficult to find a similarity measure fit for the task. In addition, substantiating why a given measure works well can be unfeasible.

\subsection{Different experimental setups}

The nature of self-training and self-training gain prediction is complex and many design choices have to be made when setting up experiments. In this section, we explore two different setups that aim at obtaining more insight in the importance of the nature of the additional data. 

A first research question is whether the additional data of the self-training in previous experiments is large enough. One may expect that more additional data would more easily lead to self-training gain.

A second experimental design change tackles the need for the similarities based on the additional data. Indeed, much information is already present in the test and training corpus. Is it useful to include extra information?

Because we change the experimental design, the number of self-training experiments is different for each change in setup. More details are given in the following paragraphs.

\paragraph{Concatenated additional data set}

In the setup of the previous self-training experiments, the unlabeled, additional data that is added is limited to the data of a single domain. The research question that is addressed in this paragraph is: Why limit the additional data to a single domain? Can the concatenated data of all domains lead to the same results?

To address this question, we set up a collection of 156 self-training experiments. The training and test corpora are the same as in the previous experiments, the additional corpus is different. The additional corpus consists of the concatenation of all 11 domains that are neither the training nor the test corpus.   

\begin{figure}[!t]
\centering
\includegraphics[scale=0.42]{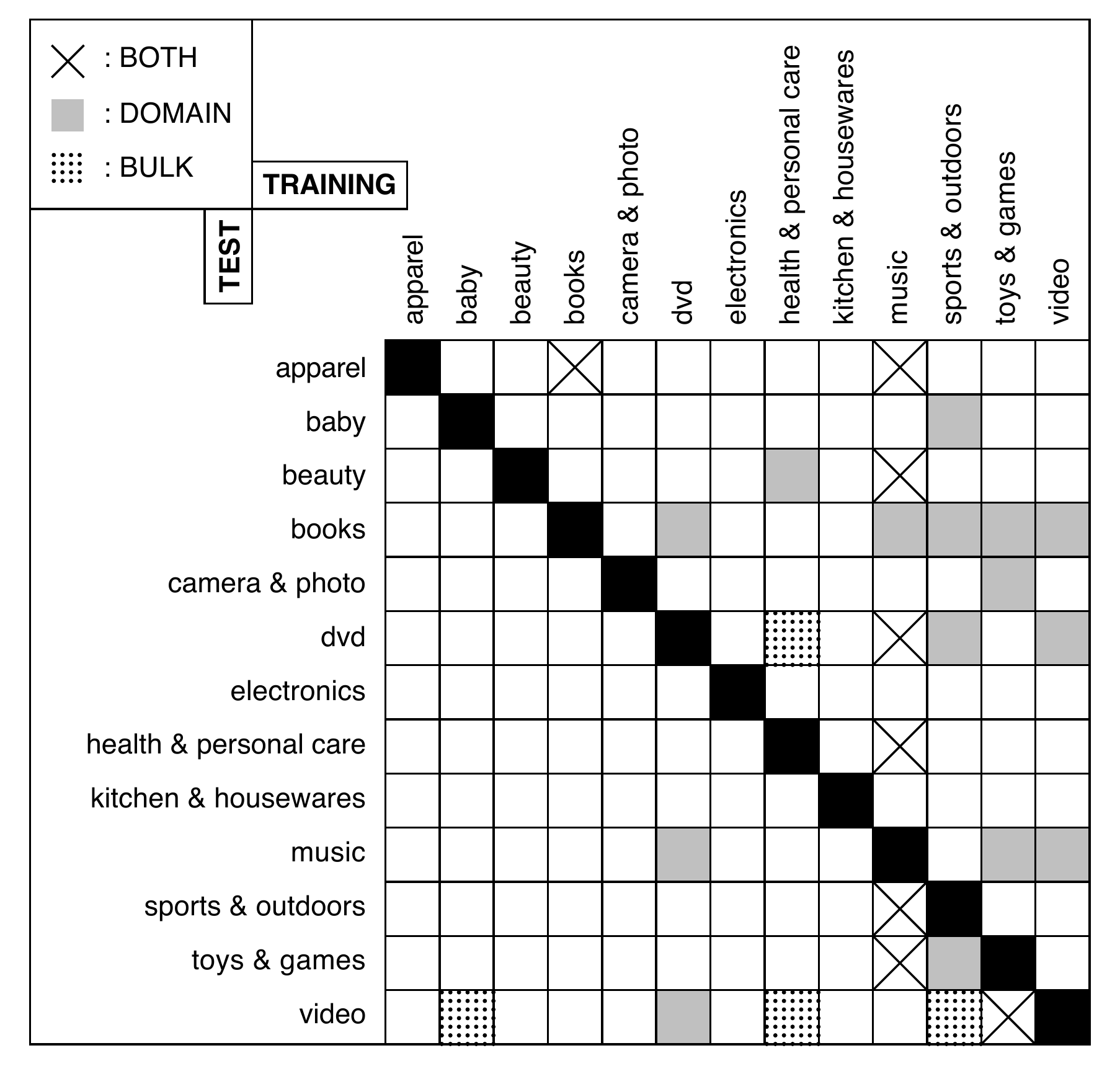}
\caption{A grid showing the self-training gain difference between a setup in which the additional data is a single domain (DOMAIN) and a setup in which the additional data is the concatenated data from all other domains (BULK). If a given training/test combination leads to self-training gain in both approaches, the cell is crossed. If only the DOMAIN approach leads to gain, the cell is greyed. If only the BULK approach leads to gain, the cell is dotted. If there is no gain, the cell is blank. Note that for the DOMAIN approach, each cell represents 11 self-training experiments and if at least one of these 11 setups leads to gain, the cell is colored. For example, using \emph{music} as training corpus and \emph{apparel} as test corpus, leads to self-training gain for both DOMAIN and BULK.}
\label{fig:grid}
\end{figure}

The overlap in setups that lead to self-training gain are given in Fig.~\ref{fig:grid}. The grid shows that there are many grey cells. These cells are the training/test combinations that lead at least once to self-training gain if a selected domain is added as unlabeled data, but there is no benefit if the concatenated data is used during self-training. This indicates that it is useful to select the right domain to add as unlabeled data, because adding all domains together eliminates the self-training gain. Since not all data is good data, it is interesting to be able to predict self-training gain. 
Fig.~\ref{fig:grid} also contains four dotted cells. These are the setups for which only the concatenated data is helpful. Including the {\footnotesize\textsc{BULK}} approach in the experiments of Section~\ref{sec:exps} would be a helpful extension to the research in this paper. We did not do this for corpus size reasons explained in Section~\ref{sec:issues}.

\paragraph{Only test/train similarity as a feature}

\begin{table}
\renewcommand{\arraystretch}{1.3}
\centering
\begin{threeparttable}[b]
\caption{Supervised performance prediction gain using tailored leave-one-out cross-validation.}
\label{tab:onlytt}
\begin{tabular}{lrrr}
\hline
type &  precision & macro-avg. & accuracy \\
 &  on gain & F-score &  \\
\hline
Cosine & 2.63 & 48.54 & 91.72 \\
Euclidean & 6.52 & 49.75 & 91.49 \\ 
KL & 42.11 & 68.03 & 92.95 \\
JS & 45.31 & 65.34 & 93.47 \\ %
sUWR & 0 & 46.61 & 87.30 \\ %
\hline
\end{tabular}
\begin{tablenotes}[flushleft]
\item Using only the test/training similarity.
\item Scores are expressed as percentages.
\end{tablenotes}
\end{threeparttable}
\end{table}

In Section~\ref{sec:exps}, the experiments are carried out using all three similarities: test/train, test/additional, and additional/train. In Section~\ref{sec:sup}, it is shown that knowledge of the the test/train combination is already very informative. For this reason, we carried out tailored leave-one-out experiments, using only the test/train similarities as a feature. The results are given in Table~\ref{tab:onlytt}. Comparison of these results with Table~\ref{tab:tlo} reveal that the scores are very similar.

The differences between the usage of one or three similarities come mainly from the fact that using three features leads to more predictions of self-training gain, i.e. self-training setups labeled as \texttt{POS}. For the Cosine distance, no extra true positives for the \texttt{POS}-class are predicted, only false positives, leading to a lower macro-averaged F-score. But in general, the extra true positives smooth out the effect of the extra false positives, leading to an increase in F-score. The only exception is the Kullback-Leibler divergence. Adding extra features leads to the prediction of less \texttt{POS}-labels. We consider this as a property of the divergence and have no explanation for this different behavior.

Based on the precision gain, using only the test/training similarity for prediction appears to be the best option. Even when the given test/training combination has previously not been seen.

\section{Conclusion}
\label{sec:con}

In this paper, we showed that self-training can be a performance boosting technique for a strict selection of setups. 

Consider a system developer who wants to implement an online labeling tool. He has one tagged corpus and a range of unlabeled corpora at his disposal. He does not know which data a user will be submitting and, as a consequence, he does not know which unlabeled corpus to add to his tagged corpus. It is also very likely that the submitted corpus will be an unseen corpus. For the binary sentiment classification data, we have shown that he could assess the similarity between his training corpus, his unlabeled corpora and the unseen test corpus and predict whether he should add the unlabeled data to his training corpus before tagging the test corpus.   

For unsupervised prediction, the best thing to do is always assume self-training loss. If one wants to be able to predict a self-training gain setup, the performance indicator can be used.

For supervised prediction, if previously computed outcomes are available for identical corpora, the best thing to do is to predict the same outcome as the majority of the identical training/test combinations. If no previous information is available for identical corpora, the similarities between the test and the training can be used to predict the outcome of the self-training experiment. The test/additional and additional/training similarities can be added as extra features, but if precision on gain is the selection criterion, this extension seems to be unnecessary.   

We do not make claims about the best choice of similarity measure because all measures have their disadvantages. Nevertheless, based on the experiments it appears that probability-distribution-based measures are more precise, and more specific: i.e.\ the Jensen-Shannon divergence.

A general conclusion regarding similarity measures is that the similarity between domains is important when evaluating domain adaptation techniques. A domain adaptation technique may lead to better results on nearby domains than on domains that are further apart. Or vice versa. Not knowing how similar domains are may lead to unjustified conclusions when comparing different domain adaptation techniques that are tested an a variety of corpora.

For this reason, a widely accepted technique to measure domain similarity would be an important addition to domain adaptation research. If further research could provide such a similarity measure, judging the applicability of domain adaptation techniques would become a lot more objective.


%

\appendices
\appendix{}
We refer to two online resources that are not essential to a clear understanding of this article:
\begin{itemize}
\item \url{www.clips.uantwerpen.be/~vincent/self-training/scripts.html}\\ For reference, the experimental implementations that are used in this study are made available online. Because the creation of a standalone application is not one of the goals of this research, the implementations are a collection of scripts rather than a mature software package.
\item \url{www.clips.uantwerpen.be/~vincent/self-training}\\ Additional insight into the structure of the corpus based on the three similarity measures as it is used in Section~\ref{sec:sup} can be gained from this vector space visualization.
\end{itemize}

\section*{Acknowledgment}
This research is funded by the Research Foundation Flanders (FWO-project G.0478.10 -- Statistical Relational Learning of Natural Language)

\ifCLASSOPTIONcaptionsoff
  \newpage
\fi



\bibliographystyle{IEEEtran}
\bibliography{IEEEabrv,bibliography}
%



%

\begin{IEEEbiographynophoto}{Vincent Van Asch}
has started his research during a project on natural language understanding, aiming to produce graphic output. After a project on biomedical text mining he obtained a PhD in Computational Linguistics at the University of Antwerp, studying the application of similarity measures on text corpora.
\end{IEEEbiographynophoto}

\begin{IEEEbiographynophoto}{Walter Daelemans}
studied linguistics and psycholinguistics at the University of Antwerp and the University of Leuven where he also received a PhD in Computational Linguistics in 1987. He held research and teaching appointments at the University of Nijmegen, the Brussels AI-Lab, and Tilburg University. Since 2005 he is full professor of Computational Linguistics at the University of Antwerp, where he also directs the CLiPS research centre. His research interests are in Computational Language Learning, Natural Language Understanding, and Computational Stylometry.
\end{IEEEbiographynophoto}





\end{document}